%% file: main.tex
\pdfoutput=1

\documentclass[11pt]{article}

\usepackage{ACL2023}

\usepackage{times}
\usepackage{latexsym}

\usepackage[T1]{fontenc}

\usepackage[utf8]{inputenc}

\usepackage{microtype}

\usepackage{inconsolata}

\usepackage[colorinlistoftodos,prependcaption,textsize=small]{todonotes}
\presetkeys{todonotes}{inline}{}

\usepackage{hyperref}
\usepackage{tabularx}
\usepackage{graphicx}
\usepackage{subfigure}
\usepackage{mathrsfs,amsmath}
\usepackage{color}
\usepackage[utf8]{inputenc}
\usepackage{amsmath,amssymb,epsfig,theorem,url,cite,bm,soul}
\usepackage{multirow}
\usepackage{wrapfig }
\usepackage{times}
\usepackage{fancyhdr,graphicx,amsmath,amssymb,enumitem}
\usepackage{tablefootnote}
\usepackage{booktabs}
\usepackage{comment}
\usepackage{mathtools}

\usepackage{xspace}
\newcommand*{\ie}{i.e.,\@\xspace}
\newcommand*{\eg}{e.g.,\@\xspace}

\newcommand{\figref}[1]{Fig.~\ref{#1}}

\newcommand{\tabref}[1]{Tab.~\ref{#1}}

\usepackage{algorithm} 
\usepackage{algorithmic}  
\usepackage[algo2e]{algorithm2e} 

\newcommand{\enes}[1]{\textcolor{purple}{E: #1}}

\usepackage{xcolor}
\def\HLred{\leavevmode\rlap{\hbox to 0.81\columnwidth{\color{red!10}\leaders\hrule height .8\baselineskip depth .5ex\hfill}}}
\def\HLblue{\leavevmode\rlap{\hbox to 0.81\columnwidth{\color{blue!10}\leaders\hrule height .8\baselineskip depth .5ex\hfill}}}
\def\HLredB{\leavevmode\rlap{\hbox to 0.88\columnwidth{\color{red!10}\leaders\hrule height .8\baselineskip depth .5ex\hfill}}}
\def\HLblueB{\leavevmode\rlap{\hbox to 0.88\columnwidth{\color{blue!10}\leaders\hrule height .8\baselineskip depth .5ex\hfill}}}

\title{Impact of Adversarial Training on Robustness and Generalizability of Language Models}

\author{
  Enes Altinisik ~ Hassan Sajjad$^{\clubsuit}$~ Husrev Taha Sencar \\  
  \textbf{Safa Messaoud} ~ \textbf{Sanjay Chawla}\\ 
{\tt \{ealtinisik,hsencar,smessaoud,schawla\}@hbku.edu.qa} \\ 
Qatar Computing Research Institute, HBKU Research Complex, Doha, Qatar \\
\\
{\tt hsajjad@dal.ca} \\ 
\textsuperscript{$\clubsuit$}Faculty of Computer Science, Dalhousie University, Halifax, Canada  \\ 
}

\begin{document}

\maketitle

\renewcommand{\headrulewidth}{0.0pt}
\thispagestyle{fancy}
\lhead{}
\rhead{}
\chead{To Appear in Findings of the Association for Computational Linguistics (ACL) 2023}
\cfoot{}

\input{abstract}
\input{introduction}
\input{related_work}

\input{label_smoothing}
\input{experiments}

\input{conclusions}

\input{limitation}

\section{Acknowledgement}
This work is partially supported by the Qatar National Research Fund (QNRF) grant NPRP11C-1229-170007.

\bibliography{bibfile}
\bibliographystyle{acl_natbib}

\end{document}

%% file: abstract.tex
\begin{abstract}
Adversarial training is widely acknowledged as the most effective defense against adversarial attacks. However, it is also well established that achieving both robustness and generalization in adversarially trained models involves a trade-off. The goal of this work is to provide an in depth comparison of different approaches for adversarial training in language models. Specifically, we study the effect of pre-training data augmentation as well as training time input perturbations vs. embedding space perturbations on the robustness and generalization of transformer-based language models. Our findings suggest that better robustness can be achieved by pre-training data augmentation or by training with input space perturbation. However, training with embedding space perturbation significantly improves generalization. A linguistic correlation analysis of neurons of the learned models reveal that the improved generalization is due to `more specialized' neurons. To the best of our knowledge, this is the first work to carry out a deep qualitative analysis of different methods of generating adversarial examples in adversarial training of language models.
\end{abstract}

%% file: introduction.tex
\section{Introduction}

Language Models (LMs) have emerged as the backbone of many tasks in AI and have extended their reach beyond NLP applications into vision and even reinforcement learning \citep{brown2020language,reed2022generalist,ramesh2022hierarchical}. Thus it is imperative that the 
generalizability and robustness of LMs be carefully assessed and evaluated.

Generalizability is the ability of a model to perform well on unseen data. Transformer-based models that are pre-trained on large unlabeled text have shown remarkable generalization ability. However, when confronted with carefully designed adversarial samples, their robustness - the ability to gracefully deal with small perturbations, 
suffers significantly. For example, a recent study has shown that on a classification task on a YELP data set, accuracy dropped by almost 90\%, when a standard test set was replaced by an adversarial counterpart \citep{jin2020bert,yoo2021towards,yuan2021bridge}.

Adversarial training is a pragmatic approach to attain both generalizability and robustness. The idea is straightforward. For a given model $M$, generate adversarial samples that target $M$ and then use the samples to incrementally re-train the model. This can be done either at the pre-training or the fine-tuning stage \citep{liu2020adversarial}.

Adversarial samples can be generated both in the input space and in the embedding space. The original work on the creation of adversarial samples for computer vision was in the input space. For example, the fast gradient sign method (FGSM)~\citep{goodfellow2014explaining} that perturbs a data point $x$ along the direction of the sign gradient of the loss function with respect to the input is an example of a perturbation in the input space. 
In the context of natural language inputs, perturbing text is challenging due to its discrete nature. 
Unlike continuous data, there is no systematical way to guarantee an increase in the loss function when perturbing text.
For instance, if we aim to make a small modification to the word ``robust'' we can choose to replace a single letter within the word or substitute it with a near synonym. 
However, both of these perturbations may seem ad-hoc and not sufficiently principled to intentionally {\em increase} the loss function. 
Therefore, in language settings, it is often more appropriate to perform perturbations in the embedding space, where continuous representations can be manipulated in a more structured manner.

Furthermore, despite the widespread use of adversarial training to increase the robustness of models, it is not clear what their impact is on downstream tasks beyond the model's overall accuracy. For example, a deeper analysis of language models has shown that different parts of the network are responsible for different parts of speech~\citep{belinkov:2017:acl,conneau2018you,liu-etal-2019-linguistic,dalvi2022discovering,durrani-etal-2020-analyzing}.
In this regard, the change in the network due to adversarial training has not yet been investigated.

Overall our contributions in this paper are three-fold.
Firstly, we introduce two techniques in the context of adversarial training in the embedding space, representing the regularization- and gradient-based approaches commonly used by latent space techniques.
We compare these techniques using a simple one-dimensional model and hypothesize their behavior in adversarial scenarios. 
Secondly, we evaluate the effectiveness of input- and embedding-space adversarial training methods in terms of their generalization ability and robustness against various types of adversarial attacks in sentiment analysis. 
Lastly, we conduct a thorough linguistic analysis of an adversarially trained model and demonstrate that incorporating robustness through adversarial training leads to more ``focused" neurons that are associated with distinct Part of Speech (POS) tags.

The rest of the paper is organized as follows.
In Section 2, we discuss adversarial attacks and defenses, with a specific focus on the NLP domain.
Section 3 provides a detailed explanation of embedding space adversarial techniques.
In Section 4, we conduct experiments to analyze the trade-off between robustness and generalization achieved by data augmentation, input-space training, and embedding space training approaches, considering various well-known adversarial attacks. Additionally, we present our findings from linguistic correlation analysis of neurons in robust models within the same section.
Finally, we finalized the paper in the concluding section.

%% file: related_work.tex
\section{Related Work}

\textbf{Adversarial Attacks: }The purpose of an adversarial attack is to cause a model to output conflicting decisions for an input and its `imperceptibly' modified version. 
An adversarial sample is defined as:
\begin{equation}
\label{eq:adversarial}
    x' = x + \delta ; ||\delta|| \le \epsilon \wedge f(x, \theta) \neq f(x', \theta)
\end{equation}
where $x'$ is the adversarial sample, $\delta$ is the perturbation added to the original data $x$,  $||\delta||$ is a generic norm, $\epsilon$ is the limit of the maximum norm of the perturbation, and $f(x, \theta)$ is the output of the model parameterized by $\theta$ for input $x$.
The quality of an adversarial sample is typically evaluated depending on how well $\delta$ is minimized, i.e., the minimum distortion that changes the prediction of the model on a sample. 

Obtaining an exact solution for the perturbation $\delta$ is a very challenging problem. 
Further, even when close approximations are considered, the solution gets computationally very expensive \citep{szegedy2013intriguing}.
To solve this problem more efficiently, gradient-based methods were introduced.
Accordingly, the perturbation $\delta$ is computed by taking one \citep{goodfellow2014explaining} or more steps iteratively \citep{madry2017towards, dong2018boosting} in the direction of the gradient to maximize the loss function.
Then, this high loss point is projected back onto the input space to determine the norm-bounded perturbation.
In practice, projected gradient descent (PGD) approaches that, take several small steps in the direction of the gradient, are used most frequently to create strong adversarial samples \citep{madry2017towards,papernot2016limitations}.

Other than gradient based approaches, \textit{Jacobian-based Saliency Map Attack} (JSMA) \citep{papernot2016limitations} uses the Jacobian matrix created from forward derivation of input to  identify to importance of each input component to the target attack.
\textit{DeepFool} \citep{moosavi2016deepfool}, alternatively, iteratively linearizes the classifier to identify the minimum perturbation that causes a change in the classification label.
\textit{Carlini \& Wagner} Attack (C\&W) proposed defensive distillation strategy \citep{hinton2015distilling} based approach.\\

\noindent\textbf{Adversarial Attacks in NLP: }Running adversarial attacks against Natural language processing (NLP) models is more challenging than widely used vision models. 
The discrete nature of word representations, combined with the tokenization of words into word pieces, effectively invalidates any algorithm that applies differential changes on the model input when generating an adversarial sample.
Moreover, quantification of the extent to which semantic similarity and contextual relations are preserved between a text input and its modified version is not trivial.

To circumvent these limitations, many adversarial sample generation algorithms adopted the approach of substituting one or more words in the input until a misprediction occurs.
The crux of this attack lies in identification of alternative words or phrases that retain the semantic intactness of the original input. 
For this, several methods based on word-embedding similarity \citep{jin2020bert}, word synonymity \citep{ren2019generating,zang2019word}, and 
masked language model predictions \citep{li2020bert} are proposed.
However, finding appropriate word candidates may get computationally very intensive.
For a sentence consisting of $m$ words with $n$ candidates to substitute each word, there are $(n + 1)^m$ possible combinations to test.
To perform this search efficiently, greedy search \citep{ren2019generating}, genetic algorithm \citep{alzantot2018generating}, and particle swarm optimization-based (PSO) \citep{zang2019word} approaches are proposed and incorporated with word importance as determined by gradient measurements \citep{yoo2021towards} and word deletion \citep{ren2019generating}.

An alternative approach to above substitution-based approach is  applying perturbations in the embedding space directly to word embeddings.
This approach avoids the expensive search step to identify the best word substitution configuration, but it requires devising a mapping 
from perturbed embeddings to the text domain in order to create an adversarial sample.
To realize this, recent work \citep{yuan2021bridge} adapted a gradient-based adversarial sample generation method to compute perturbations associated with each word embedding.
Perturbed embeddings are then translated to input domain using a pre-trained masked-language modeling (MLM) head, as in \citep{li2020bert,garg2020bae}, to create an adversarial sample that is semantically similar to the original input. \\

\noindent\textbf{Adversarial Defence in NLP: }The most commonly deployed method for attaining robustness against an adversarial attack is through addition of adversarial samples into the training set \citep{szegedy2013intriguing}.
This approach is known to increase model robustness in both computer vision and NLP domains.
Further, it is also reported that this defence approach decreases the generalization error of a model in the absence of any attack \citep{yuan2021bridge}, which contradicts the commonly held opinion that there is a trade-off between generalization and robustness \enes{\citep{tsipras2018robustness}}.
This finding can essentially be attributed to the use of a larger training set enhanced with adversarial samples.
The second approach augments the training set with newly constructed, synthetic samples.
While this may seem equivalent to adding adversarial samples to the training set, data augmentation methods do not need to have an adversarial nature.
Common data augmentation methods include word replacement, i.e., substituting words with their synonyms or inserting random words, random word deletions, and swapping of words between sentences \citep{wei2019eda}.
Rather than using manually-designed heuristics, the power of existing NLP models can also be harnessed for data augmentation. 
Reverse translation, which involves re-translation of samples from a target language back to their source language constitutes one such method that ideally preserves the semantic similarity of original and augmented samples \citep{edunov2018understanding,xie2020unsupervised}.  
The use of MLM via masking words in a sentence and replacing them with model predictions \citep{ng2020ssmba} is another augmentation method.

The third approach to adversarial training involves applying perturbations in the latent space \citep{zhu2019freelb,liu2020adversarial,li2021token,pan2022improved}.
This yields a simpler training procedure as it removes the need for generating adversarial samples in the input space. 
In \citep{zhu2019freelb}, a model is incrementally fine-tuned on sets of adversarially perturbed word embeddings computed after each fine-tuning step.
\citet{li2021searching} demonstrate that this method performs better when no constraint on the amount of perturbation is imposed.
In \citet{li2021token}, it is observed that rather than initializing the PGD step with random noise when computing perturbations for each token, using a token-dependent random noise that is fixed across all inputs is more effective. 
Recently, \citet{pan2022improved} proposed the use of contrastive objective \citep{oord2018representation} for ensuring invariant representations by forcing the model to learn the differences between the normal input and its adversarial version.

In addition to empirical methods, certified defense methods are proposed to identify and eliminate adversarial samples. 
These techniques minimize misclassification within an $l_{\infty}$ ball bound, particularly in the vision domain \citep{raghunathan2018certified, wong2018provable}. 
In the NLP domain, two main categories of certified defense methods have emerged: Interval Bound Propagation (IBP) \citep{jia2019certified, huang2019achieving, Shi2020Robustness} and randomized smoothing \citep{ye2020safer, zeng2021certified}.
IBP techniques estimate the output range by iteratively applying interval constraints from the input layer to subsequent layers.
However, the requirement to modify the model structure poses challenges in incorporating these methods into pre-trained models.

Randomized smoothing-based methods offer an alternative approach that is independent of the model structure. 
These methods utilize stochastic ensembles of input texts and leverage the statistical properties of these ensembles to offer provable robustness certification. 
A common approach to achieve this is by generating a few randomly modified versions of the original sample. 
This can be done through techniques such as random word substitutions using synonyms, as demonstrated in SAFER \citep{ye2020safer}, or by employing a mask language model to substitute words, as shown in RanMASK \citep{zeng2021certified}. 
The final prediction is then made based on the decisions made by these randomly generated samples.

Throughout the rest of the paper, we do not delve into a detailed discussion of these techniques for several reasons. 
Firstly, the main focus of this paper is on empirical methods and evaluating their impact. 
Secondly, randomized smoothing methods can be integrated into various techniques, making them applicable in different contexts. 
Lastly, previous findings suggest that while randomized smoothing methods demonstrate strong defense performance, they tend to underperform compared to latent space adversarial training \citep{li2021searching}.

%% file: label_smoothing.tex
\section{AT with Embedding Space Perturbations}

Among all adversarial defenses developed for language processing models, moving the adversarial training from the input space to the embedding space offers the most advantage.
This essentially allows the adoption of gradient-based adversarial training approaches that are computationally less demanding than input space methods. 
Although a plethora of such adversarial training methods exists, they are all essentially guided by two main principles in their approach. 
The first one essentially sets the training objective to minimize the loss due to worst-case perturbation induced on the training samples, instead of the average loss computed from training samples by the standard training. 
This group of methods essentially differ in the way they approximate the worst-case perturbation \citep{madry2017towards,miyato2018virtual,zhang2019theoretically} as well as the extent and nature of perturbation applied during generation of adversarial samples \citep{ding2018mma,wang2019improving, liu2020adversarial}.

The second approach primarily relies on the premise that smoothness is an important requirement of a robust model. 
To this objective, these methods focus on minimization of a regularized version of the loss instead of optimizing only the standard, training loss. 
The regularization term here ensures that there is a wide enough margin around each training data point with the decision boundary of the model 
through minimizing the difference between the predictions of natural and adversarial samples.
Methods following this approach are distinguished based on their formulation of regularization \citep{szegedy2016rethinking,zhang2019theoretically} and their coupling with the training loss described above \citep{villa,pan2022improved}.

In our analysis, we consider two representative methods that most effectively exemplify each approach.
In practice, due to its computational efficiency, the PGD attack is most frequently used for the creation of adversarial samples. 
We will refer to this generic adversarial training approach as PGD-AT.
The latter approach is also best characterized by the use of PGD in ensuring local distribution smoothness around natural samples. 
This alternative method will be referred to as LDS. 
We must note that improved variants of the two base methods should be expected to perform better.
In this regard, robustness-generalization performance of the PGD-AT and LDS can be interpreted as lower-bounds.

\begin{algorithm}[!tb]
\small
\caption{PGD-AT and LDS based adversarial training}
\label{alg:alternativeAlg}
	\textbf{Input:}  $E$: the number of epochs, $D=\{(x_{(i)}, y_{(i)})\}_{i=1}^n$:  the dataset,
	$f(x, \theta)$: the machine learning model parametrized by $\theta$, $\delta$: the perturbation initialized by $\sigma$ and limited by $\epsilon$, $\tau$: the global learning rate, 
	$\mu$: the adversarial learning rate, $S$: the number of PGD step, and $\Pi$ is the projection function. \\
	\For{$e=1,..,E$}{
        \For{$(x,y) \in \mathcal{D}$}{
            $\delta \sim \mathcal{N}(0,\sigma^{2})$\\
            \For{$s=1,..,S$}{
                 \HLred $g_{adv}= \nabla_{\delta} l(f(x+\delta,\theta),y)$          \%PGD-AT \\
                 \HLblue $g_{adv}= \nabla_{\delta} l(f(x,\theta),f(x+\delta, \theta))$  \%LDS \\
    	         $\delta = \Pi_{||\delta|| \leq \epsilon} (\delta+\mu g_{adv})$
	        }
	         \HLredB  $g_{\theta} \leftarrow \nabla_{\theta} l(f(x, \theta), y)$\\
	             \HLredB \hskip2em $+ \nabla_{\theta} l(f(x+\delta, \theta), y) $ \%PGD-AT \\
	         \HLblueB  $g_{\theta} \leftarrow \nabla_{\theta} l(f(x, \theta), y)$\\
    	         \HLblueB \hskip2em $+ \nabla_{\theta} l(f(x,\theta),f(x+\delta, \theta)) $ \%LDS \\
	         $\theta \leftarrow \theta - \tau g_{\theta}$
        }
    }	
	\textbf{Output:}  $\theta$
\end{algorithm}

The steps of both methods are presented in Algorithm \ref{alg:alternativeAlg}
where the lines that differ between the two methods are highlighted as pink for PGD-AT and blue for LDS.
Both methods start by randomly initializing $\delta$ with normal distribution with a mean of zero and standard deviation of $\sigma$. 
The loss is then calculated between the model's output of the perturbed input depending on the method, PGD-AT or LDS. 
The $\delta$ value is then updated by the gradient and clipped to within $\pm \epsilon$ by the projection function $\Pi$.
These steps are repeated for $S$ times. 
The loss value is then updated by combining the standard loss with the loss associated with each method. 
Gradient update is then applied to model parameters. 

\begin{table}
\small
\resizebox{\linewidth}{!}{
\begin{tabular}{|c|c|c|} \hline 
Model     & Loss Function & Parameter \\ \hline \hline
   OLS  & $\frac{1}{n}\sum_{i=1}^{n}(\theta.x_{i} - y_{i})^2$ & $\theta = \frac{\sum_{i}x_{i}y_{i}}{\sum_{i}x_{i}^2}$ \\ \hline
   \multirow{2}{*}{PGD-AT}  & $\frac{1}{n}\sum_{i=1}^{n}\left\{(\theta.x_{i} - y_{i})^2\right. + $ & \multirow{2}{*}{$\theta = \frac{\sum_{i}2x_{i}y_{i} + y_{i}\delta}{\sum_{i}x_{i}^2 + (x_{i} + \delta)^2}$} \\ 
   & $\left.(\theta.(x_{i} + \delta) - y_{i})^2\right\}$ &   \\  \hline
   \multirow{2}{*}{LDS}  & $\frac{1}{n}\sum_{i=1}^{n}\left\{(\theta.x_{i} - y_{i})^2\right. + $ & \multirow{2}{*}{$\theta = \frac{\sum_{i}x_{i}y_{i}}{\sum_{i}x_{i}^2 + \delta^2}$} \\ 
   & $\left.(\theta.(x_{i} + \delta) - \theta.x_{i})^2\right\}$ &   \\  \hline
   \end{tabular}
}
\caption{Closed form solutions of the model parameter of a one-dimensional linear regression model under various loss functions}
\label{table:formulation}
\end{table}

\begin{figure}[htbp]
  \begin{minipage}[b]{0.49\columnwidth}
    \centering
    \includegraphics[width=1\columnwidth]{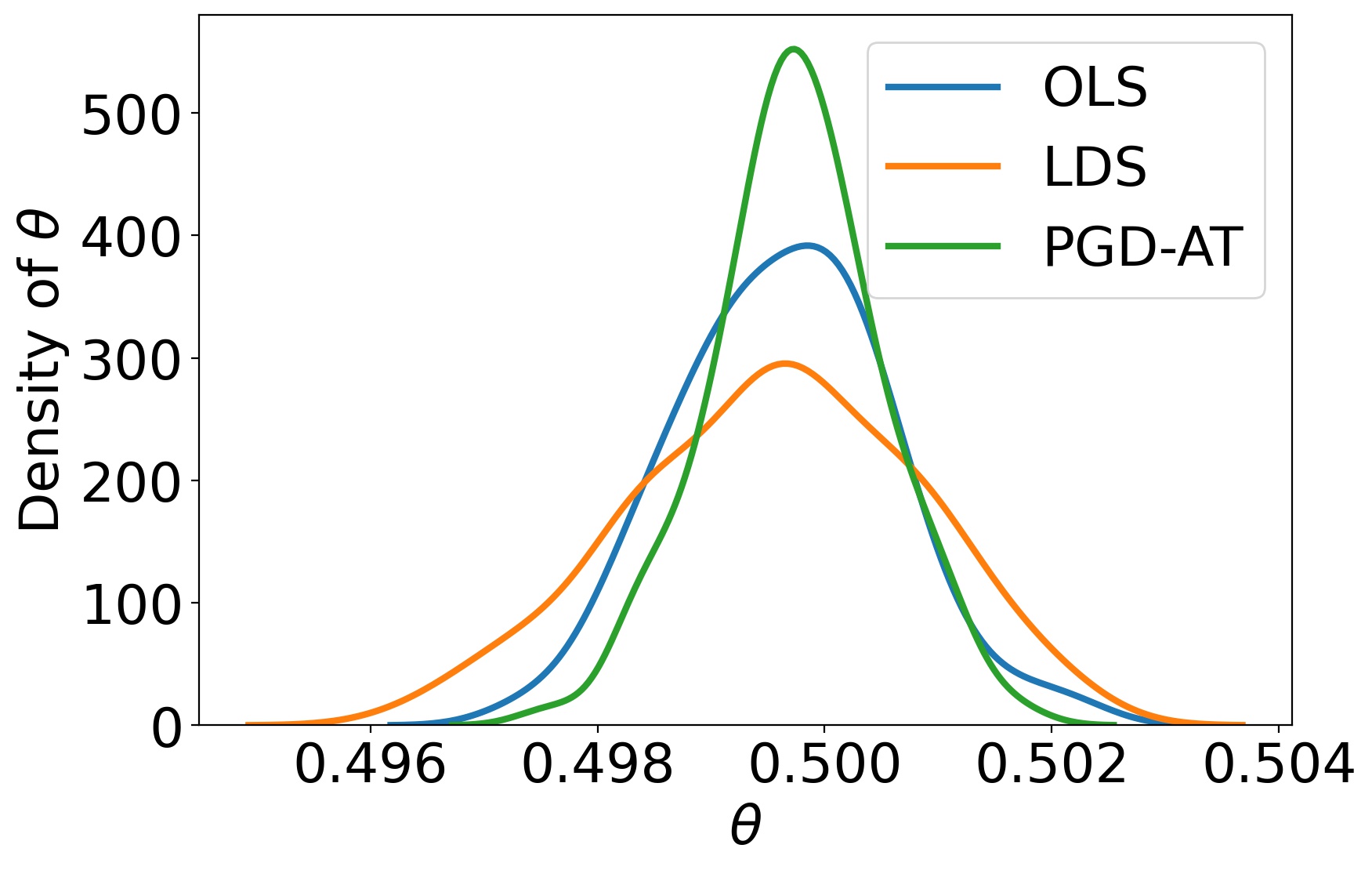}
    \centerline{\scriptsize{(a)}}
  \end{minipage}
  \begin{minipage}[b]{0.49\columnwidth}
    \centering
    \includegraphics[width=1\columnwidth]{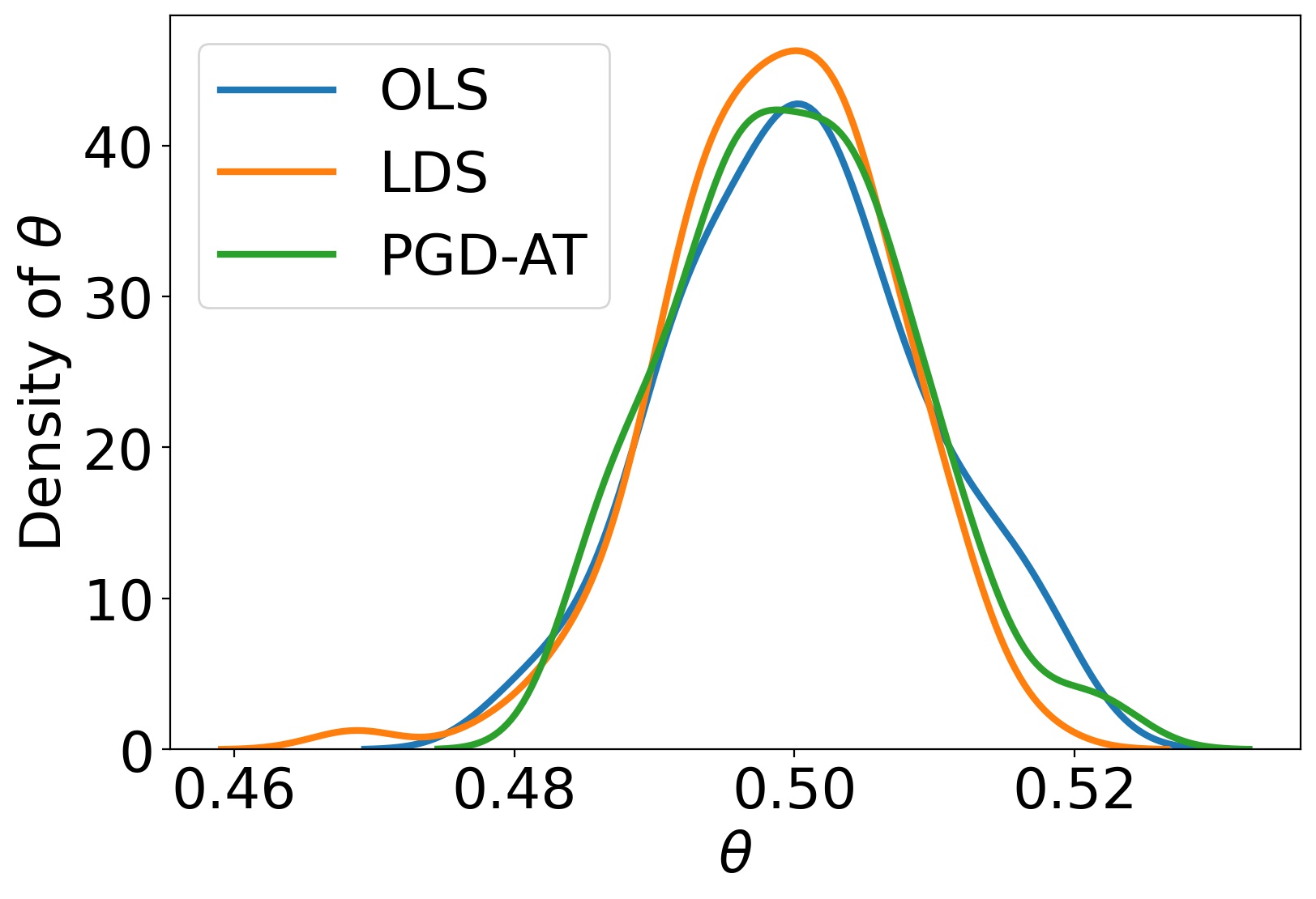}
    \centerline{\scriptsize{(b)}}
  \end{minipage}
  \caption{The resulting distribution for $\theta$ values related to three different models, trained using OLS, LDS, and PGD-AT methods, when $\sigma$ is set to (a) 0.01 and (b) 0.1. 
  A small standard deviation indicates the model's robustness and clustering around 0.5 implies better generalizability.}
  \label{fig:liner}
\end{figure}

\input{Figs_sections/fig_evaluation}
To better examine the behavior of the two methods, we analyze a simple one-dimensional linear regression model:
\[y = \theta.x + \epsilon, \hspace{1cm} \epsilon \sim N(0,\sigma^2)
\]
Assuming a fixed perturbation $\delta$, we determine how the two loss functions, given in Algorithm 1, estimate the model parameter $\theta$ under noisy observations.
Table~\ref{table:formulation} presents the loss functions corresponding to PGD-AT and LDS as well as the one corresponding to the standard ordinary least squares (OLS) estimation in the absence of $\delta$.
The estimates for the parameter $\theta$ for the three loss functions are also given in the table (third column). 
Comparing PGD-AT and LDS, it can be deduced that LDS will converge to OLS only as the noise $\epsilon$ gets severe, suppressing the effect of $\delta$ in the denominator.
Whereas PGD-AT can be expected to follow OLS more closely at all noise levels as $\delta$ appears both at the numerator and the denominator, thereby absorbing its effect on the estimate.

We also designed an experimental setup to test these hypotheses. 
A single neuron is trained based on randomly generated ($x,y$) pairs as defined above assuming $\theta=\frac{1}{2}$ and for two different noise distributions, ($\sigma=0.01$ and $\sigma=0.1$) for each loss function.
The models are trained for 2K epochs at a learning rate of 0.005 starting with the OLS loss.  
For PGD-AT and LDS models, the OLS loss is substituted by their loss function after epoch 1750 and $\delta$ values are computed as defined in Algorithm 1.

The distributions of the estimated scalar model parameter $\theta$ obtained after 25 runs is displayed in Fig. \ref{fig:liner}.
Essentially, the spread of the distribution signifies the robustness of a model against adversarial samples and the distribution mean relates to the generalizability of the model. 
In this regard, PGD-AT is seen to perform better than LDS as it yields a tighter spread in both cases. 
However, at higher noise levels, it can be seen that LDS provides a more accurate estimate of $\theta$.
Overall, we can expect that a model trained with PGD-AT to be more robust while yielding a generalizability behavior closer to that of LDS.

%% file: Figs_sections/fig_evaluation.tex
\begin{figure}[!h]
    \centering
    \includegraphics[width=1.0\columnwidth]{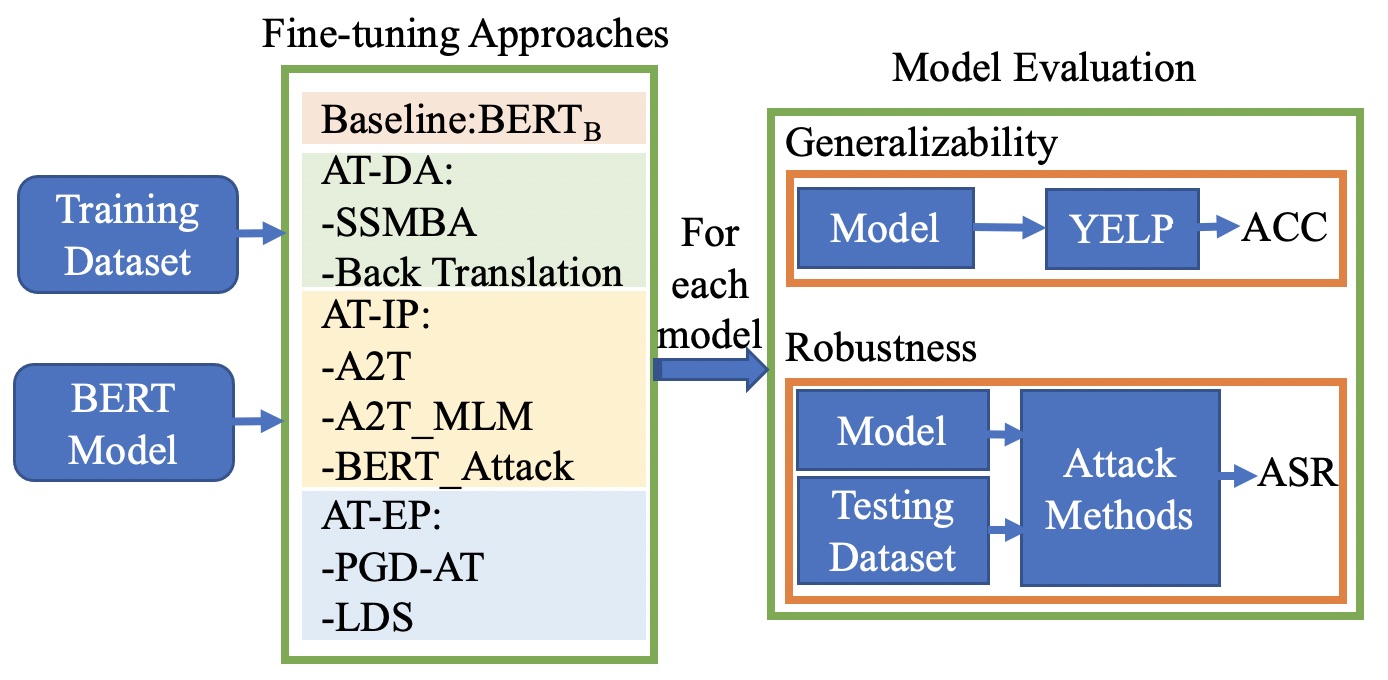}
  \caption{Evaluation pipeline of models learned using different adversarial training approaches.}
  \label{fig:eval_pipeline}
\end{figure}

\begin{figure*}[!h]
    \centering
    \includegraphics[width=1.6\columnwidth]{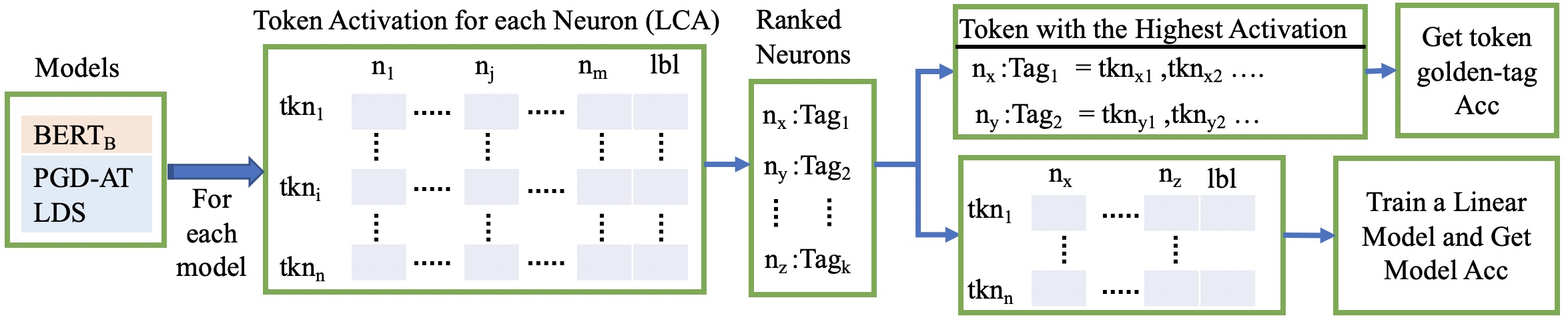}
  \caption{LCA pipeline of models learned using different adversarial training approaches.}
  \label{fig:_pipeline}
\end{figure*}

%% file: experiments.tex
\input{Tables_sections/tab_robustness} 
\section{Experiments}

We first compare the robustness, generalization and run-time complexity of different AT strategies, following the pipeline in \figref{fig:eval_pipeline}. Then, we perform a Linguistic Correlation Analysis ~\citep[LCA, ][]{dalvi:2019:AAAI} as implemented in the NeuroX toolkit~\citep{neurox-acl23:demo} to gain better insights into the dynamics of the learned models, as illustrated in \figref{fig:_pipeline}.
\\
\noindent\textbf{Baselines:} We compare standard BERT \citep{devlin2018bert} with seven versions of adversarially trained BERT models using methods from three families of AT approaches: (1) AT with pre-training data augmentation (AT-DA), (2) AT with input space perturbations (AT-IP) and (3) AT with embedding space perturbations (AT-EP), on the task of sentiment classification. Specifically, for AT-DA, we experiment with SSMBA \citep{ng2020ssmba} and BackTranslation \citep{xie2020unsupervised}. For AT-IP, we use A2T, A2T\_MLM \citep{yoo2021towards} and BERT\_attack \citep{li2020bert}. For AT-EP, we report results on LDS \citep{szegedy2016rethinking,zhang2019theoretically} and PGD-AT \citep{villa,pan2022improved}. \\
\textbf{Datasets}: We fine-tune all models on the Internet Movie Database ~\citep[IMDB, ][]{maas2011learning} and Movie Reviews ~\citep[MR, ][]{pang2005seeing} datasets and test on the corresponding testing splits, as well as on YELP dataset \citep{zhang2015character} for  out-of-distribution assessment of the models.\\
\textbf{Attack methods:} We assess the robustness of the models under four different attacks which replace words in the input space using different strategies. (1) TextFooler \citep{jin2020bert} first searches for the word that results in the highest change in the sentiment score, when removed, then replaces it with the nearest neighbouring word in the embedding space. (2) BAE \citep{garg2020bae} masks a portion of the text and using a BERT masked language model to generate alternatives for the masked words. (3) A2T \citep{yoo2021towards} selects the word with the largest loss gradient w.r.t its embedding and replaces it with a synonym generated from a counterfitted word embedding \citep{mrkvsic2016counter}. (4) PSO \citep{zang2019word} uses sememe-based word substitution and particle swarm optimization-based
search algorithm to find good adversarial examples for a given input text.\\
\textbf{Evaluation metrics:} we assess (1) generalization via computing the accuracy values on in-distribution and out-of-distribution datasets, (2) robustness using the Attack Success Rate (ASR) representing the ratio of the number of successful attacks to the number of samples, as well as (3) the time complexity measured via the fine-tuning run-time of the BERT model over 4 epochs.\\
\textbf{Implementation details:} For AT-DA and AT-IP methods, we use the parameters proposed by the corresponding papers. For our PGD-AT and LDS approaches, we limit the number of PGD steps to 3 and the perturbations L2-norm to $0.003$. All experiments are conducted on Nvidia v100 Tensor Core GPU.\\
\noindent\textbf{Run-time results:} We report the time for fine-tuning the models over 4 epochs in \tabref{table:runTime}. The AT-DA approaches results in the shortest fine-tuning time as adversarial examples are generated once for every sample before the training, unlike in AT-IP and AT-EP where adversarial examples are generated at every training iteration. AT-EP methods, are around 1.5 times slower to fine-tune than the standard BERT model as generating the adversarial examples requires an additional backward pass for computing the gradient of the loss, at every training iteration. As expected, AT-IP methods are the most time consuming as they involve a combinatorial search over a large number of input space configurations. For example, the fastest approach in this class, A2T, needs 6 seconds for a single adversarial example generation, which is around 10 times slower than the other approaches.
\input{Tables_sections/tab_run_time}

\noindent\textbf{Robustness results} are shown in \tabref{table:robustness}. The lower the ASR the better is the model in withstanding the attack. As expected, the most effective methods against adversarial attacks are the AT-IP ones. This is due to the fact that the only class of approaches were it's possible to match the attack and the defense strategies, \ie train on perturbations generated from the attack strategies, is AT-IP, as attacks in language models operate in the input space. Among AT-AD methods, BackTranslation is the most robust method on the IMDB dataset. We found that this is due to IMDB having in average long sentences which makes it easier to generate good and diverse adversarial examples to train on, via back translation. Our results show that AT-EP methods are the least robust. In particles, LDS-AT struggle in the sentiment classification task due to noisy ground-truth label, \ie sentiments are mostly not binary but the ground truth labels are.\\
\textbf{Generalization results} are reported in \tabref{table:accuracy}. AT-DA accuracy values are comparable to BERT. Hence, it looks like AT-DA generalization capabilities are not traded-off for better robustness as it is the case of AT-IP approaches. This is due to the fact that adversarial examples from SSMBA (self-supervised-based) and BackTranslation (translation-based) are generated while taking the global context into account. So they are unlikely to change the semantics of the input text and hence the decision boundaries. These methods are however unpractical for usage inside of the training loop. More efficient techniques, \eg based on local search in the embedding space, are used by AT-IP methods. This however might not always lead to preserving the semantics of the original input text, which also means that assigning the label of the ground truth input to these adversarial examples might be inappropriate or noisy. Such hard examples are well known to encourage overfitting and hence reduce the generalization ability of the model. This explains the significant drop in both in and out-of-distribution accuracy values of AT-IP approaches. The best generalization results are obtained using AT-EP methods. We notice that PGD-AT consistently improves upon BERT. This phenomena doesn't occur in vision where generalization is well know to drop in adversarially trained models. To the best of our knowledge, we are the first to report this in language models trained with embedding space perturbation. In order to gain a better understanding of the reasons behind this phenomena, we investigate the learned dynamics of deepnets trained with AT-EP methods using Linguistic Correlation Analysis (next paragraph). Specifically, we want to validate that the achieved accuracy was due to better learning to solve of the task at hand and not just due to memorizing the training data. 
\input{Tables_sections/tab_generalization}

\input{Tables_sections/tab_lca_1}

\input{appendix.tex}

\input{Tables_sections/tab_lca_2} 

\noindent\textbf{Linguistic Correlation Analysis ~\citep[LCA, ][]{dalvi:2019:AAAI}} is used to identify the most salient neurons for a given linguistic
property like a Parts-of-Speech (POS) tag~\citep{sajjad-etal-2022-neuron}. To achieve this, we first match words to neurons, then assess if the matched words have the linguistic property of interest. As the sentiment prediction task is not appropriate for word level analysis, \ie same words can be part of different sentiment classes, we focus on POS tagging task. We fine-tune BERT models using AT-EP methods on the publicly available Penn Treebank dataset \citep{marcinkiewicz1994building}. We use LCA to generate a list of the top-5 firing neurons for every POS tag and leverage these lists to perform two types of analysis: (1) neurons-POS tags association strength analysis and a (2) a neural ablation analysis. To assess the neurons-tag association strength, given the list of the top-firing neurons from LCA, we next generate a list of the words in the testing data with the highest activation values for these neurons. Then, we compute the intersection between the generated word list and the ground-truth one, \ie the list of words with label being the POS tag of interest in the testing data. A large intersection set means that the neurons learned to specialize in predicting specific POS tags, \ie they learned the linguistic nuances of the task and are unlikely to have just memorized the training data. Results in \tabref{tab:POSdist}\footnote{Definitions of POS tags with their order in the table: adjective; adjective, comparative; modal; verb, past tense; colon, semi-colon; verb, 3rd person singular present; adverb; verb, gerund or present participle} show that our AT-EP learn more `focused' neurons as measured by the intersection ratio (match/total). In particular, PGD-AT significantly improves upon the standard BERT$_\text{B}$ model. 

Table \ref{tab:word2POS} provides words corresponding to select POS tags obtained from the models trained with the $BERT_B$, the LDS, and PGD-AT methods.

\noindent For the second analysis, \ie the neural ablation study, we create a linear regression model using only activations of the top 10 ranked neurons. Results are shown in \tabref{tab:lds_2}. PGD-AT and LDA achieve a significantly higher performance than BERT, which further support the observation that AT helped better learn the intricacies of the tasks and explains the improvement of the generalization abilities of the AT-EP approaches (\eg in \tabref{table:accuracy}).

%% file: Tables_sections/tab_robustness.tex
\begin{table*}[t]
    \centering
    \caption{Robustness results. Models are evaluated using ASR (lower is better) on the MR and IMDB datasets.}
    \vspace{-0.2cm}
    \resizebox{\linewidth}{!}{
    \begin{tabular}{cccccccccc}
    \toprule
    \multirow{2}{*}{Attack}     & \multirow{2}{*}{Dataset} & \multirow{2}{*}{BERT} & \multicolumn{3}{c}{AT-IP} & \multicolumn{2}{c}{AT-DA} & \multicolumn{2}{c}{AT-EP} \\
    \cmidrule(lr){4-6} \cmidrule(lr){7-8} \cmidrule(lr){9-10} 
                                &         &          &A2T &A2T\_MLM &BERT\_Attack &SSMBA & BackTranslation & LDS & PGD-AT\\  
    \midrule
    \multirow{2}{*}{TextFooler} & MR      & 82.1	& 77.9	& 79.7	& 79.3	& 79.8	& \textbf{72.8}	& 88.6	& 87.8 \\
                                & IMDB    & 80.6	& 86.5	& 65.3	& 72.8	& 81.3	& \textbf{45.9}	& 91.5	& 94.7 \\ \hline

    \multirow{2}{*}{A2T}        & MR      &  33.8	& 27.6	& 30.8	& 26.5	& 34.2	& 30.4	& 22.3	& \textbf{20.9} \\ 
                                & IMDB    &  59.5	& 51.1	& 43.7	& 49.4	& 59.2	& \textbf{36.4}	& 56.9	& 43.0 \\ \hline

    \multirow{2}{*}{BAE}        & MR      & 52.1	& 44.1	& 45.5	& \textbf{44.0}	& 49.2	& 47.1	& 55.3	& 52.9 \\ 
                                & IMDB    & 68.8	& 65.0	& 52.5	& 57.9	& 61.5	& \textbf{41.4}	& 66.5	& 61.0  \\ \hline
                                
    \multirow{2}{*}{PSO}        & MR      & 79.8	& 75.0	& \textbf{72.7}	& 74.7	& 78.1	& 75.6	& 79.8	& 80.7 \\ 
                                & IMDB    & 46.4	& 35.3	& 35.4	& \textbf{30.2}	& 41.8	& 42.8	& 70.8	& 66.2 \\   \midrule 
                                
    \multirow{2}{*}{\textbf{Average}}& MR & 62.0	& \textbf{56.1}	& 57.2	& \textbf{56.1}	& 60.3	& 56.5	& 61.5	&  60.5 \\ 
                                & IMDB    & 63.8	& 59.5	& 49.2	& 52.6	& 61.0	& \textbf{41.6}	& 71.4	&  66.2 \\ 
    \bottomrule
    \end{tabular}
    }
    \label{table:robustness}
\end{table*}

%% file: Tables_sections/tab_run_time.tex
\begin{table}[h!]
    \centering
    \caption{ Run-time results. We report the fine-tuning run-time over 4 episodes on the MR and IMDB datasets.}
    \vspace{-0.2cm}
    \renewcommand{\arraystretch}{1.35}
    \setlength\tabcolsep{13pt}
    \small{
    \begin{tabular}{lccc}
    \toprule
    &\multirow{2}{*}{ Models } & \multicolumn{2}{c}{Run Time (in min)}\\
    \cmidrule{3-4}
    &&\begin{tabular}[c]{@{}c@{}}IMDB\end{tabular} & \begin{tabular}[c]{@{}c@{}}MR\end{tabular}   \\ \hline
    
    & BERT                    & \textbf{79.0}	& \textbf{38.2}  \\  \hline
    \multirow{2}{*}{\rotatebox[origin=c]{90}{AT-DA}}& SSMBA             & 112.8	&46.4  \\  
    & BackTranslation       & 210.5	&66.0  \\  \hline
    \multirow{3}{*}{\rotatebox[origin=c]{90}{AT-IP}}& A2T               & 1600.5	&448.5  \\  
    & A2T\_MLM          & 1494.3	&504.7  \\  
    & BERT\_Attack      & 1495.2	&461.5   \\  \hline
    \multirow{2}{*}{\rotatebox[origin=c]{90}{AT-EP}}& LDS             & 163.4  &64.2   \\  
    & PGD-AT          & 158.2  &69.0   \\  
    \bottomrule
    \end{tabular}
    }
    \label{table:runTime}
\end{table}

%% file: Tables_sections/tab_generalization.tex
\begin{table}[h!]
    \centering
    \caption{Generalization results. We report the accuracy values on IMDB/MR (in-distribution) and YELP (out-of-distribution) datasets for BERT models fine-tuned on IMDB/MR for the task of sentiment classification. }
    \vspace{-0.2cm}
    \renewcommand{\arraystretch}{1.35}
    \resizebox{\columnwidth}{!}{
    \begin{tabular}{lccccc}
    \toprule
    &\multirow{3}{*}{ Models } & \multicolumn{2}{c}{IMDB} & \multicolumn{2}{c}{MR}\\
    \cmidrule(lr){3-4} \cmidrule(lr){5-6} 
    &&\begin{tabular}[c]{@{}c@{}}IMDB\end{tabular} & \begin{tabular}[c]{@{}c@{}}YELP\end{tabular}  &
    \begin{tabular}[c]{@{}c@{}}MR\end{tabular} &
    \begin{tabular}[c]{@{}c@{}}YELP\end{tabular} \\  
    \midrule
    &BERT     & 93.49	& 91.24 	& 85.27	& 87.06  \\  \hline 
    \multirow{2}{*}{\rotatebox[origin=c]{90}{AT-DA}} &SSMBA    & 93.49	& 91.17     & 85.24	& 87.72  \\
    &BackTranslation     & 93.44	& 91.50	& 84.96	 & 87.77  \\  \hline
    \multirow{3}{*}{\rotatebox[origin=c]{90}{AT-IP}}&A2T               & 92.59 & 89.97 & 83.58 & 83.62  \\  
    &A2T\_MLM         & 92.70 & 89.15 & 83.90 & 81.79  \\  
    &BERT\_Attack      & 92.63 & 90.04 & 84.61& 80.41  \\  \hline
    \multirow{2}{*}{\rotatebox[origin=c]{90}{AT-EP}}&LDS              & 93.24 & \textbf{92.09}  & 86.49& 81.80    \\ 
    &PGD-AT         & \textbf{93.80} & \textbf{92.11} & \textbf{86.59}        & \textbf{88.16} \\   \bottomrule
    \end{tabular}
    }
    \label{table:accuracy}
\end{table}

%% file: Tables_sections/tab_lca_1.tex
\begin{table}[!h]
	\centering
	\caption{LCA results. The association strength between POS tags and neurons.} 
	\vspace{-0.2cm}
	\resizebox{\linewidth}{!}{
    	\begin{tabular}{cccccccccc}
    	\toprule
    	    \multirow{2}{*}{ POS } & \multicolumn{3}{c}{BERT} & \multicolumn{3}{c}{LDS} & \multicolumn{3}{c}{PGD-AT}\\
            \cmidrule(lr){2-4} \cmidrule(lr){5-7} \cmidrule(lr){8-10} 
            &\begin{tabular}[c]{@{}c@{}}Match\end{tabular} & \begin{tabular}[c]{@{}c@{}}Total\end{tabular}  &
            \begin{tabular}[c]{@{}c@{}}\%\end{tabular}  &
            \begin{tabular}[c]{@{}c@{}}Match\end{tabular} & \begin{tabular}[c]{@{}c@{}}Total\end{tabular}&
            \begin{tabular}[c]{@{}c@{}}\%\end{tabular}  &
            \begin{tabular}[c]{@{}c@{}}Match\end{tabular} & \begin{tabular}[c]{@{}c@{}}Total\end{tabular}&
            \begin{tabular}[c]{@{}c@{}}\%\end{tabular}  \\ 
            \midrule
                JJ	&2	&15&13.33	&2	&15&13.33	&\textbf{6}	&10	& \textbf{60.00}\\ 
                JJR	&3	&9&	33.33&4	&9	&44.44&\textbf{7}&13	&\textbf{53.84}\\ 
                MD	&0	&5&	0.00&\textbf{3}	&5&\textbf{60.00}	&2	&5&	40.00\\ 
                VBD	&\textbf{5}	&5& \textbf{100.00}&0	&5&	0.00&0	&5&	0.00\\ 
                :	&0	&5& 0.00	&1	&5	& 20.00 &\textbf{3}	&5&	\textbf{60.00}\\
                VBZ	&4	&10&40.00	&7	&9	&77.77&\textbf{9}	&10	&\textbf{90.00}\\ 
                RB	&\textbf{9}	&10&\textbf{90.00}	&\textbf{9}	&10&\textbf{90.00}	&6	&10	&60.00\\
                VBG	&10	&12&83.33	&14	&18	&77.77&\textbf{15} &15& \textbf{100.00}\\
    	\bottomrule
    	\end{tabular}
	}

	\label{tab:POSdist}
\end{table}

%% file: appendix.tex
\begin{table*}[!t]
	\centering
	\caption{Examples of the most related words for different POS tags for models trained with the $BERT_B$, the LDS, and PGD-AT methods. The words are bolded when their actual tags match with the associated tag, where the actual tags correspond to the most frequent tags of the words based on the POS-tagged training data.}
	\vspace{-0.2cm}
	\resizebox{\linewidth}{!}{
    	\begin{tabular}{c|c|c|c}
    	\toprule
        POS                 & $BERT_B$                     & LDS               & PGD-AT\\ \hline
        
       \multirow{3}{*}{VBZ}  & \textbf{indicates} teenage And \textbf{begins} 	& \textbf{indicates} \textbf{denies} \textbf{erodes} \textbf{explains} 	& \textbf{indicates} \textbf{accounts} \textbf{refuses} \textbf{agrees} 		\\
                            & \textbf{reflects} \textbf{explains} evil Previously 	&  \textbf{resembles} And \textbf{runs} 	& \textbf{is} \textbf{has} \textbf{believes} And 		\\
                            & automatic reckless 	& \textbf{adds} trains 	& \textbf{adds} \textbf{begins}  \\
                             \hline
                            
       \multirow{3}{*}{JJ}  & Rae away \textbf{little} Springs Nelson	& Aktiebolaget least plummeted Do policies	& \textbf{bright} away what \textbf{high}   \\
                            & live \textbf{equal} What explain Giants	& \textbf{little} told What \textbf{equal} securities	& \textbf{strong} \textbf{cold} skyrocketed  \\ 
                            & Who Aktiebolaget skyrocketed what rung	& Dallara added said most cardboard	& \textbf{green} What \textbf{same} \\
                             \hline
                             
       \multirow{3}{*}{JJR}  & \textbf{newer} meaning \textbf{greater} punish 	& included \textbf{newer} \textbf{greater} \textbf{smaller} 	& included \textbf{newer} \textbf{stronger} meaning  \\
                            & included banking close 	& indicated shipbuilding arranged 	& \textbf{smaller} \textbf{greater} indicated planning   \\ 
                            & \textbf{smaller}  her 	& \textbf{Higher}  her 	& \textbf{higher} close \textbf{Higher} \textbf{lower} least\\
                             \hline
                            
        \multirow{1}{*}{MD} &  associated bright required  severe denied		& apart \textbf{shall} \textbf{might} \textbf{must}  fallen	& fallen \textbf{shall} expected \textbf{might} apart 	\\ 
                             \hline
                            
        \multirow{1}{*}{VBD}& \textbf{restored} \textbf{bothered} \textbf{notched}	\textbf{mixed} \textbf{began}  & expire face exist  become buy 	& expire face become  exist disagree	\\
                  
    \bottomrule
    	\end{tabular}
	}
	
	\label{tab:word2POS}
\end{table*}

%% file: Tables_sections/tab_lca_2.tex
\begin{table}[!h]
\centering
\caption{LCA results. Neural ablation study.}
\vspace{-0.2cm}
\setlength\tabcolsep{22pt}
\small{
\begin{tabular}{ccccc} 
\toprule
BERT & LDS  & PGD-AT   \\ \midrule
34.2\% &  38.6\% & 35.3\%   \\
\bottomrule
\end{tabular}}
\label{tab:lds_2}
\end{table}

%% file: conclusions.tex
\section{Conclusions}
In this paper we have carried out an extensive study of adversarial training methods (ATMs) to understand their
impact on robustness and generalizability for transformer-based deep language models. 
We can draw the following conclusions from our study. First, non-adversarial data augmentation improves both generalization and robustness over the baseline BERT model. 
Adversarial training in the input space yields better robustness compared to both non-adversarial data augmentation and embedding space adversarial training. 
In contrast, adversarial training in the embedding space exhibits best generalization behavior. 
Among PGD-AT and LDS methods, our results show that the PGD-AT is consistently more robust and generalizable.
Overall, our results show that unlike in computer vision domain where gradient-based adversarial training yields the best robustness and generalization trade-off, for language processing models input-space training methods are indispensable.

For future work we will consider combining data augmentation, input-space training, and embedding space training approaches together.
We would also like to extend our theoretical understanding of the trade-off between robustness and generalizability for language models. 
In connection, the impact of ATMs for other downstream applications needs to be studied.

%% file: limitation.tex
\section*{Limitations}
All our experiments are performed using the BERT-small language model due to the computational requirements of generating and testing models considering many configurations of  adversarial training and attack methods. 
Although using larger language models might have provided different performance measurements, our findings that compare input- and embedding-space adversarial training methods are expected to remain unchanged. Another limitation of our work is the semantic gap between attacks in input and embedding space needs further research. Specifically,  how do perturbations in the embedding space get translated in the input space? Finally, other forms of robustness techniques, besides adversarial training, in the context of large language models require examination.

\section*{Ethics Statement}
The work studied the impact of several adversarial training methods on robustness and generalization. The work did not result in any new dataset and model and it has no potential ethical issues. On the positive side, the work targets two important attributes of trustworthy AI i.e. robustness and generalization. Our work provides an insightful comparison of the input-space and embedding space adversarial training approaches and will positively impact the future research work in this area.

%% file: main.bbl
\begin{thebibliography}{56}
\expandafter\ifx\csname natexlab\endcsname\relax\def\natexlab#1{#1}\fi

\bibitem[{Alzantot et~al.(2018)Alzantot, Sharma, Elgohary, Ho, Srivastava, and
  Chang}]{alzantot2018generating}
Moustafa Alzantot, Yash Sharma, Ahmed Elgohary, Bo-Jhang Ho, Mani Srivastava,
  and Kai-Wei Chang. 2018.
\newblock Generating natural language adversarial examples.
\newblock \emph{arXiv preprint arXiv:1804.07998}.

\bibitem[{Belinkov et~al.(2017)Belinkov, Durrani, Dalvi, Sajjad, and
  Glass}]{belinkov:2017:acl}
Yonatan Belinkov, Nadir Durrani, Fahim Dalvi, Hassan Sajjad, and James Glass.
  2017.
\newblock \href
  {https://aclanthology.coli.uni-saarland.de/pdf/P/P17/P17-1080.pdf} {{What do
  Neural Machine Translation Models Learn about Morphology?}}
\newblock In \emph{Proceedings of the 55th Annual Meeting of the Association
  for Computational Linguistics (ACL)}, Vancouver. Association for
  Computational Linguistics.

\bibitem[{Brown et~al.(2020)Brown, Mann, Ryder, Subbiah, Kaplan, Dhariwal,
  Neelakantan, Shyam, Sastry, Askell et~al.}]{brown2020language}
Tom Brown, Benjamin Mann, Nick Ryder, Melanie Subbiah, Jared~D Kaplan, Prafulla
  Dhariwal, Arvind Neelakantan, Pranav Shyam, Girish Sastry, Amanda Askell,
  et~al. 2020.
\newblock Language models are few-shot learners.
\newblock \emph{Advances in neural information processing systems},
  33:1877--1901.

\bibitem[{Conneau et~al.(2018)Conneau, Kruszewski, Lample, Barrault, and
  Baroni}]{conneau2018you}
Alexis Conneau, German Kruszewski, Guillaume Lample, Lo{\"\i}c Barrault, and
  Marco Baroni. 2018.
\newblock {What you can cram into a single vector: Probing sentence embeddings
  for linguistic properties}.
\newblock In \emph{Proceedings of the 56th Annual Meeting of the Association
  for Computational Linguistics (ACL)}.

\bibitem[{Dalvi et~al.(2019)Dalvi, Durrani, Sajjad, Belinkov, Bau, and
  Glass}]{dalvi:2019:AAAI}
Fahim Dalvi, Nadir Durrani, Hassan Sajjad, Yonatan Belinkov, D.~Anthony Bau,
  and James Glass. 2019.
\newblock What is one grain of sand in the desert? analyzing individual neurons
  in deep nlp models.
\newblock In \emph{Proceedings of the Thirty-Third AAAI Conference on
  Artificial Intelligence (AAAI, Oral presentation)}.

\bibitem[{Dalvi et~al.(2022)Dalvi, Khan, Alam, Durrani, Xu, and
  Sajjad}]{dalvi2022discovering}
Fahim Dalvi, Abdul~Rafae Khan, Firoj Alam, Nadir Durrani, Jia Xu, and Hassan
  Sajjad. 2022.
\newblock \href {https://openreview.net/forum?id=POTMtpYI1xH} {Discovering
  latent concepts learned in {BERT}}.
\newblock In \emph{International Conference on Learning Representations}.

\bibitem[{Dalvi et~al.(2023)Dalvi, Sajjad, and Durrani}]{neurox-acl23:demo}
Fahim Dalvi, Hassan Sajjad, and Nadir Durrani. 2023.
\newblock Neurox library for neuron analysis of deep nlp models.
\newblock In \emph{Proceedings of the Association for Computational Linguistics
  (ACL)}.

\bibitem[{Devlin et~al.(2018)Devlin, Chang, Lee, and
  Toutanova}]{devlin2018bert}
Jacob Devlin, Ming-Wei Chang, Kenton Lee, and Kristina Toutanova. 2018.
\newblock Bert: Pre-training of deep bidirectional transformers for language
  understanding.
\newblock \emph{arXiv preprint arXiv:1810.04805}.

\bibitem[{Ding et~al.(2018)Ding, Sharma, Lui, and Huang}]{ding2018mma}
Gavin~Weiguang Ding, Yash Sharma, Kry Yik~Chau Lui, and Ruitong Huang. 2018.
\newblock Mma training: Direct input space margin maximization through
  adversarial training.
\newblock \emph{arXiv preprint arXiv:1812.02637}.

\bibitem[{Dong et~al.(2018)Dong, Liao, Pang, Su, Zhu, Hu, and
  Li}]{dong2018boosting}
Yinpeng Dong, Fangzhou Liao, Tianyu Pang, Hang Su, Jun Zhu, Xiaolin Hu, and
  Jianguo Li. 2018.
\newblock Boosting adversarial attacks with momentum.
\newblock In \emph{Proceedings of the IEEE conference on computer vision and
  pattern recognition}, pages 9185--9193.

\bibitem[{Durrani et~al.(2020)Durrani, Sajjad, Dalvi, and
  Belinkov}]{durrani-etal-2020-analyzing}
Nadir Durrani, Hassan Sajjad, Fahim Dalvi, and Yonatan Belinkov. 2020.
\newblock \href {https://doi.org/10.18653/v1/2020.emnlp-main.395} {Analyzing
  individual neurons in pre-trained language models}.
\newblock In \emph{Proceedings of the 2020 Conference on Empirical Methods in
  Natural Language Processing (EMNLP)}, pages 4865--4880, Online. Association
  for Computational Linguistics.

\bibitem[{Edunov et~al.(2018)Edunov, Ott, Auli, and
  Grangier}]{edunov2018understanding}
Sergey Edunov, Myle Ott, Michael Auli, and David Grangier. 2018.
\newblock Understanding back-translation at scale.
\newblock \emph{arXiv preprint arXiv:1808.09381}.

\bibitem[{Gan et~al.(2020)Gan, Chen, Li, Zhu, Cheng, and Liu}]{villa}
Zhe Gan, Yen-Chun Chen, Linjie Li, Chen Zhu, Yu~Cheng, and Jingjing Liu. 2020.
\newblock Large-scale adversarial training for vision-and-language
  representation learning.
\newblock \emph{Advances in Neural Information Processing Systems},
  33:6616--6628.

\bibitem[{Garg and Ramakrishnan(2020)}]{garg2020bae}
Siddhant Garg and Goutham Ramakrishnan. 2020.
\newblock Bae: Bert-based adversarial examples for text classification.
\newblock \emph{arXiv preprint arXiv:2004.01970}.

\bibitem[{Goodfellow et~al.(2014)Goodfellow, Shlens, and
  Szegedy}]{goodfellow2014explaining}
Ian~J Goodfellow, Jonathon Shlens, and Christian Szegedy. 2014.
\newblock Explaining and harnessing adversarial examples.
\newblock \emph{arXiv preprint arXiv:1412.6572}.

\bibitem[{Hinton et~al.(2015)Hinton, Vinyals, Dean
  et~al.}]{hinton2015distilling}
Geoffrey Hinton, Oriol Vinyals, Jeff Dean, et~al. 2015.
\newblock Distilling the knowledge in a neural network.
\newblock \emph{arXiv preprint arXiv:1503.02531}, 2(7).

\bibitem[{Huang et~al.(2019)Huang, Stanforth, Welbl, Dyer, Yogatama, Gowal,
  Dvijotham, and Kohli}]{huang2019achieving}
Po-Sen Huang, Robert Stanforth, Johannes Welbl, Chris Dyer, Dani Yogatama, Sven
  Gowal, Krishnamurthy Dvijotham, and Pushmeet Kohli. 2019.
\newblock Achieving verified robustness to symbol substitutions via interval
  bound propagation.
\newblock \emph{arXiv preprint arXiv:1909.01492}.

\bibitem[{Jia et~al.(2019)Jia, Raghunathan, G{\"o}ksel, and
  Liang}]{jia2019certified}
Robin Jia, Aditi Raghunathan, Kerem G{\"o}ksel, and Percy Liang. 2019.
\newblock Certified robustness to adversarial word substitutions.
\newblock \emph{arXiv preprint arXiv:1909.00986}.

\bibitem[{Jin et~al.(2020)Jin, Jin, Zhou, and Szolovits}]{jin2020bert}
Di~Jin, Zhijing Jin, Joey~Tianyi Zhou, and Peter Szolovits. 2020.
\newblock Is bert really robust? a strong baseline for natural language attack
  on text classification and entailment.
\newblock In \emph{Proceedings of the AAAI conference on artificial
  intelligence}, volume~34, pages 8018--8025.

\bibitem[{Li et~al.(2020)Li, Ma, Guo, Xue, and Qiu}]{li2020bert}
Linyang Li, Ruotian Ma, Qipeng Guo, Xiangyang Xue, and Xipeng Qiu. 2020.
\newblock Bert-attack: Adversarial attack against bert using bert.
\newblock \emph{arXiv preprint arXiv:2004.09984}.

\bibitem[{Li and Qiu(2021)}]{li2021token}
Linyang Li and Xipeng Qiu. 2021.
\newblock Token-aware virtual adversarial training in natural language
  understanding.
\newblock In \emph{Proceedings of the AAAI Conference on Artificial
  Intelligence}, volume~35, pages 8410--8418.

\bibitem[{Li et~al.(2021)Li, Xu, Zeng, Li, Zheng, Zhang, Chang, and
  Hsieh}]{li2021searching}
Zongyi Li, Jianhan Xu, Jiehang Zeng, Linyang Li, Xiaoqing Zheng, Qi~Zhang,
  Kai-Wei Chang, and Cho-Jui Hsieh. 2021.
\newblock Searching for an effective defender: Benchmarking defense against
  adversarial word substitution.
\newblock \emph{arXiv preprint arXiv:2108.12777}.

\bibitem[{Liu et~al.(2019)Liu, Gardner, Belinkov, Peters, and
  Smith}]{liu-etal-2019-linguistic}
Nelson~F. Liu, Matt Gardner, Yonatan Belinkov, Matthew~E. Peters, and Noah~A.
  Smith. 2019.
\newblock \href {https://www.aclweb.org/anthology/N19-1112} {Linguistic
  knowledge and transferability of contextual representations}.
\newblock In \emph{Proceedings of the 2019 Conference of the North {A}merican
  Chapter of the Association for Computational Linguistics: Human Language
  Technologies, Volume 1 (Long and Short Papers)}, pages 1073--1094,
  Minneapolis, Minnesota. Association for Computational Linguistics.

\bibitem[{Liu et~al.(2020)Liu, Cheng, He, Chen, Wang, Poon, and
  Gao}]{liu2020adversarial}
Xiaodong Liu, Hao Cheng, Pengcheng He, Weizhu Chen, Yu~Wang, Hoifung Poon, and
  Jianfeng Gao. 2020.
\newblock Adversarial training for large neural language models.
\newblock \emph{arXiv preprint arXiv:2004.08994}.

\bibitem[{Maas et~al.(2011)Maas, Daly, Pham, Huang, Ng, and
  Potts}]{maas2011learning}
Andrew Maas, Raymond~E Daly, Peter~T Pham, Dan Huang, Andrew~Y Ng, and
  Christopher Potts. 2011.
\newblock Learning word vectors for sentiment analysis.
\newblock In \emph{Proceedings of the 49th annual meeting of the association
  for computational linguistics: Human language technologies}, pages 142--150.

\bibitem[{Madry et~al.(2017)Madry, Makelov, Schmidt, Tsipras, and
  Vladu}]{madry2017towards}
Aleksander Madry, Aleksandar Makelov, Ludwig Schmidt, Dimitris Tsipras, and
  Adrian Vladu. 2017.
\newblock Towards deep learning models resistant to adversarial attacks.
\newblock \emph{arXiv preprint arXiv:1706.06083}.

\bibitem[{Marcinkiewicz(1994)}]{marcinkiewicz1994building}
Mary~Ann Marcinkiewicz. 1994.
\newblock Building a large annotated corpus of english: The penn treebank.
\newblock \emph{Using Large Corpora}, 273.

\bibitem[{Miyato et~al.(2018)Miyato, Maeda, Koyama, and
  Ishii}]{miyato2018virtual}
Takeru Miyato, Shin-ichi Maeda, Masanori Koyama, and Shin Ishii. 2018.
\newblock Virtual adversarial training: a regularization method for supervised
  and semi-supervised learning.
\newblock \emph{IEEE transactions on pattern analysis and machine
  intelligence}, 41(8):1979--1993.

\bibitem[{Moosavi-Dezfooli et~al.(2016)Moosavi-Dezfooli, Fawzi, and
  Frossard}]{moosavi2016deepfool}
Seyed-Mohsen Moosavi-Dezfooli, Alhussein Fawzi, and Pascal Frossard. 2016.
\newblock Deepfool: a simple and accurate method to fool deep neural networks.
\newblock In \emph{Proceedings of the IEEE conference on computer vision and
  pattern recognition}, pages 2574--2582.

\bibitem[{Mrk{\v{s}}i{\'c} et~al.(2016)Mrk{\v{s}}i{\'c}, S{\'e}aghdha, Thomson,
  Ga{\v{s}}i{\'c}, Rojas-Barahona, Su, Vandyke, Wen, and
  Young}]{mrkvsic2016counter}
Nikola Mrk{\v{s}}i{\'c}, Diarmuid~O S{\'e}aghdha, Blaise Thomson, Milica
  Ga{\v{s}}i{\'c}, Lina Rojas-Barahona, Pei-Hao Su, David Vandyke, Tsung-Hsien
  Wen, and Steve Young. 2016.
\newblock Counter-fitting word vectors to linguistic constraints.
\newblock \emph{arXiv preprint arXiv:1603.00892}.

\bibitem[{Ng et~al.(2020)Ng, Cho, and Ghassemi}]{ng2020ssmba}
Nathan Ng, Kyunghyun Cho, and Marzyeh Ghassemi. 2020.
\newblock Ssmba: Self-supervised manifold based data augmentation for improving
  out-of-domain robustness.
\newblock \emph{arXiv preprint arXiv:2009.10195}.

\bibitem[{Oord et~al.(2018)Oord, Li, and Vinyals}]{oord2018representation}
Aaron van~den Oord, Yazhe Li, and Oriol Vinyals. 2018.
\newblock Representation learning with contrastive predictive coding.
\newblock \emph{arXiv preprint arXiv:1807.03748}.

\bibitem[{Pan et~al.(2022)Pan, Hang, Sil, and Potdar}]{pan2022improved}
Lin Pan, Chung-Wei Hang, Avirup Sil, and Saloni Potdar. 2022.
\newblock Improved text classification via contrastive adversarial training.
\newblock In \emph{Proceedings of the AAAI Conference on Artificial
  Intelligence}, volume~36, pages 11130--11138.

\bibitem[{Pang and Lee(2005)}]{pang2005seeing}
Bo~Pang and Lillian Lee. 2005.
\newblock Seeing stars: Exploiting class relationships for sentiment
  categorization with respect to rating scales.
\newblock \emph{arXiv preprint cs/0506075}.

\bibitem[{Papernot et~al.(2016)Papernot, McDaniel, Jha, Fredrikson, Celik, and
  Swami}]{papernot2016limitations}
Nicolas Papernot, Patrick McDaniel, Somesh Jha, Matt Fredrikson, Z~Berkay
  Celik, and Ananthram Swami. 2016.
\newblock The limitations of deep learning in adversarial settings.
\newblock In \emph{2016 IEEE European symposium on security and privacy
  (EuroS\&P)}, pages 372--387. IEEE.

\bibitem[{Raghunathan et~al.(2018)Raghunathan, Steinhardt, and
  Liang}]{raghunathan2018certified}
Aditi Raghunathan, Jacob Steinhardt, and Percy Liang. 2018.
\newblock \href {https://openreview.net/forum?id=Bys4ob-Rb} {Certified defenses
  against adversarial examples}.
\newblock In \emph{International Conference on Learning Representations}.

\bibitem[{Ramesh et~al.(2022)Ramesh, Dhariwal, Nichol, Chu, and
  Chen}]{ramesh2022hierarchical}
Aditya Ramesh, Prafulla Dhariwal, Alex Nichol, Casey Chu, and Mark Chen. 2022.
\newblock Hierarchical text-conditional image generation with clip latents.
\newblock \emph{arXiv preprint arXiv:2204.06125}.

\bibitem[{Reed et~al.(2022)Reed, Zolna, Parisotto, Colmenarejo, Novikov,
  Barth-Maron, Gimenez, Sulsky, Kay, Springenberg et~al.}]{reed2022generalist}
Scott Reed, Konrad Zolna, Emilio Parisotto, Sergio~Gomez Colmenarejo, Alexander
  Novikov, Gabriel Barth-Maron, Mai Gimenez, Yury Sulsky, Jackie Kay,
  Jost~Tobias Springenberg, et~al. 2022.
\newblock A generalist agent.
\newblock \emph{arXiv preprint arXiv:2205.06175}.

\bibitem[{Ren et~al.(2019)Ren, Deng, He, and Che}]{ren2019generating}
Shuhuai Ren, Yihe Deng, Kun He, and Wanxiang Che. 2019.
\newblock Generating natural language adversarial examples through probability
  weighted word saliency.
\newblock In \emph{Proceedings of the 57th annual meeting of the association
  for computational linguistics}, pages 1085--1097.

\bibitem[{Sajjad et~al.(2022)Sajjad, Durrani, and
  Dalvi}]{sajjad-etal-2022-neuron}
Hassan Sajjad, Nadir Durrani, and Fahim Dalvi. 2022.
\newblock \href {https://doi.org/10.1162/tacl_a_00519} {Neuron-level
  interpretation of deep {NLP} models: A survey}.
\newblock \emph{Transactions of the Association for Computational Linguistics},
  10:1285--1303.

\bibitem[{Shi et~al.(2020)Shi, Zhang, Chang, Huang, and
  Hsieh}]{Shi2020Robustness}
Zhouxing Shi, Huan Zhang, Kai-Wei Chang, Minlie Huang, and Cho-Jui Hsieh. 2020.
\newblock \href {https://openreview.net/forum?id=BJxwPJHFwS} {Robustness
  verification for transformers}.
\newblock In \emph{International Conference on Learning Representations}.

\bibitem[{Szegedy et~al.(2016)Szegedy, Vanhoucke, Ioffe, Shlens, and
  Wojna}]{szegedy2016rethinking}
Christian Szegedy, Vincent Vanhoucke, Sergey Ioffe, Jon Shlens, and Zbigniew
  Wojna. 2016.
\newblock Rethinking the inception architecture for computer vision.
\newblock In \emph{Proceedings of the IEEE conference on computer vision and
  pattern recognition}, pages 2818--2826.

\bibitem[{Szegedy et~al.(2013)Szegedy, Zaremba, Sutskever, Bruna, Erhan,
  Goodfellow, and Fergus}]{szegedy2013intriguing}
Christian Szegedy, Wojciech Zaremba, Ilya Sutskever, Joan Bruna, Dumitru Erhan,
  Ian Goodfellow, and Rob Fergus. 2013.
\newblock Intriguing properties of neural networks.
\newblock \emph{arXiv preprint arXiv:1312.6199}.

\bibitem[{Tsipras et~al.(2019)Tsipras, Santurkar, Engstrom, Turner, and
  Madry}]{tsipras2018robustness}
Dimitris Tsipras, Shibani Santurkar, Logan Engstrom, Alexander Turner, and
  Aleksander Madry. 2019.
\newblock \href {https://openreview.net/forum?id=SyxAb30cY7} {Robustness may be
  at odds with accuracy}.
\newblock In \emph{International Conference on Learning Representations}.

\bibitem[{Wang et~al.(2019)Wang, Zou, Yi, Bailey, Ma, and
  Gu}]{wang2019improving}
Yisen Wang, Difan Zou, Jinfeng Yi, James Bailey, Xingjun Ma, and Quanquan Gu.
  2019.
\newblock Improving adversarial robustness requires revisiting misclassified
  examples.
\newblock In \emph{International Conference on Learning Representations}.

\bibitem[{Wei and Zou(2019)}]{wei2019eda}
Jason Wei and Kai Zou. 2019.
\newblock Eda: Easy data augmentation techniques for boosting performance on
  text classification tasks.
\newblock \emph{arXiv preprint arXiv:1901.11196}.

\bibitem[{Wong and Kolter(2018)}]{wong2018provable}
Eric Wong and Zico Kolter. 2018.
\newblock Provable defenses against adversarial examples via the convex outer
  adversarial polytope.
\newblock In \emph{International conference on machine learning}, pages
  5286--5295. PMLR.

\bibitem[{Xie et~al.(2020)Xie, Dai, Hovy, Luong, and Le}]{xie2020unsupervised}
Qizhe Xie, Zihang Dai, Eduard Hovy, Thang Luong, and Quoc Le. 2020.
\newblock Unsupervised data augmentation for consistency training.
\newblock \emph{Advances in Neural Information Processing Systems},
  33:6256--6268.

\bibitem[{Ye et~al.(2020)Ye, Gong, and Liu}]{ye2020safer}
Mao Ye, Chengyue Gong, and Qiang Liu. 2020.
\newblock Safer: A structure-free approach for certified robustness to
  adversarial word substitutions.
\newblock \emph{arXiv preprint arXiv:2005.14424}.

\bibitem[{Yoo and Qi(2021)}]{yoo2021towards}
Jin~Yong Yoo and Yanjun Qi. 2021.
\newblock Towards improving adversarial training of nlp models.
\newblock \emph{arXiv preprint arXiv:2109.00544}.

\bibitem[{Yuan et~al.(2021)Yuan, Zhang, Chen, and Wei}]{yuan2021bridge}
Lifan Yuan, Yichi Zhang, Yangyi Chen, and Wei Wei. 2021.
\newblock Bridge the gap between cv and nlp! a gradient-based textual
  adversarial attack framework.
\newblock \emph{arXiv preprint arXiv:2110.15317}.

\bibitem[{Zang et~al.(2019)Zang, Qi, Yang, Liu, Zhang, Liu, and
  Sun}]{zang2019word}
Yuan Zang, Fanchao Qi, Chenghao Yang, Zhiyuan Liu, Meng Zhang, Qun Liu, and
  Maosong Sun. 2019.
\newblock Word-level textual adversarial attacking as combinatorial
  optimization.
\newblock \emph{arXiv preprint arXiv:1910.12196}.

\bibitem[{Zeng et~al.(2021)Zeng, Zheng, Xu, Li, Yuan, and
  Huang}]{zeng2021certified}
Jiehang Zeng, Xiaoqing Zheng, Jianhan Xu, Linyang Li, Liping Yuan, and Xuanjing
  Huang. 2021.
\newblock Certified robustness to text adversarial attacks by randomized
  [mask].
\newblock \emph{arXiv preprint arXiv:2105.03743}.

\bibitem[{Zhang et~al.(2019)Zhang, Yu, Jiao, Xing, El~Ghaoui, and
  Jordan}]{zhang2019theoretically}
Hongyang Zhang, Yaodong Yu, Jiantao Jiao, Eric Xing, Laurent El~Ghaoui, and
  Michael Jordan. 2019.
\newblock Theoretically principled trade-off between robustness and accuracy.
\newblock In \emph{International conference on machine learning}, pages
  7472--7482. PMLR.

\bibitem[{Zhang et~al.(2015)Zhang, Zhao, and LeCun}]{zhang2015character}
Xiang Zhang, Junbo Zhao, and Yann LeCun. 2015.
\newblock Character-level convolutional networks for text classification.
\newblock \emph{Advances in neural information processing systems}, 28.

\bibitem[{Zhu et~al.(2019)Zhu, Cheng, Gan, Sun, Goldstein, and
  Liu}]{zhu2019freelb}
Chen Zhu, Yu~Cheng, Zhe Gan, Siqi Sun, Tom Goldstein, and Jingjing Liu. 2019.
\newblock Freelb: Enhanced adversarial training for language understanding.

\end{thebibliography}
